\documentclass{article} 
\usepackage{iclr2025_conference,times}

\usepackage{amsmath,amsfonts,bm}

\def\eqref#1{equation~\ref{#1}}

\def\1{\bm{1}}

\DeclareMathAlphabet{\mathsfit}{\encodingdefault}{\sfdefault}{m}{sl}
\SetMathAlphabet{\mathsfit}{bold}{\encodingdefault}{\sfdefault}{bx}{n}

\usepackage{hyperref}
\usepackage{url}

\usepackage{microtype}
\usepackage{graphicx}
\usepackage{subfigure}
\usepackage{booktabs} 
\usepackage{mdframed}

\usepackage{amsmath}
\usepackage{amssymb}
\usepackage{mathtools}
\usepackage{amsthm}

\usepackage{enumitem}
\usepackage{dsfont}
\usepackage{tikz, pgfplots}
\usepackage{multirow}
\usepackage{adjustbox}
\usepackage{float}
\usepackage{multicol}
\usepackage{wrapfig}
\usepackage[ruled, vlined]{algorithm2e}
\usepackage{algpseudocode}

\usetikzlibrary{positioning, arrows.meta}
\definecolor{lever_green}{HTML}{4C9A53}
\definecolor{medium_red}{RGB}{220,0,0}
\definecolor{dark_red}{RGB}{244,30,20}
\definecolor{dark_yellow}{RGB}{180,180,10}
\definecolor{dark_green}{RGB}{0, 110, 0}
\definecolor{dark_blue}{RGB}{35, 13, 248}
\definecolor{dark_orange}{RGB}{204, 130, 0}
\definecolor{dark_purple}{RGB}{100, 0, 255}
\definecolor{dark_pink}{RGB}{219, 100, 150}

\newcommand{\bfres}[2]{\bfseries{#1} $\pm$ \bfseries{#2}}

\mdfsetup{skipabove=0pt,skipbelow=0pt}

\title{Symmetry-Breaking Augmentations for Ad Hoc Teamwork}

\author{Ravi Hammond\textsuperscript{\textnormal{ 1 2}}, Dustin Craggs\textsuperscript{\textnormal{ 2}}, Mingyu Guo\textsuperscript{\textnormal{ 2}}, Jakob Foerster\textsuperscript{\textnormal{ 1 3}} \& Ian Reid\textsuperscript{\textnormal{ 2 4}}\\
\textsuperscript{\textnormal{1}}Foerster Lab for AI Research, University of Oxford\\
\textsuperscript{\textnormal{2}}Australian Institute for Machine Learning, University of Adelaide\\
\textsuperscript{\textnormal{3}}Meta AI Research, UK\\
\textsuperscript{\textnormal{4}}Mohamed bin Zayed University of
Artificial Intelligence\\
\texttt{ravihammond@gmail.com}
}

\iclrfinalcopy 
\pgfplotsset{compat=1.18}
\begin{document}

\maketitle

\begin{abstract}
In dynamic collaborative settings, for artificial intelligence (AI) agents to better align with humans, they must adapt to novel teammates who utilise unforeseen strategies. While adaptation is often simple for humans, it can be challenging for AI agents. Our work introduces \textit{symmetry-breaking augmentations} (SBA) as a novel approach to this challenge. By applying a symmetry-flipping operation to increase \textit{behavioural diversity} among training teammates, SBA encourages agents to learn robust responses to unknown strategies, highlighting how social conventions impact human-AI alignment. We demonstrate this experimentally in two settings, showing that our approach outperforms previous ad hoc teamwork results in the challenging card game Hanabi. In addition, we propose a general metric for estimating symmetry dependency amongst a given set of policies. Our findings provide insights into how AI systems can better adapt to diverse human conventions and the core mechanics of alignment.
\end{abstract}

\begin{wrapfigure}{r}{0.5\textwidth} 
  \centering
  \vspace{-40pt}
  \includegraphics[width=0.48\textwidth]{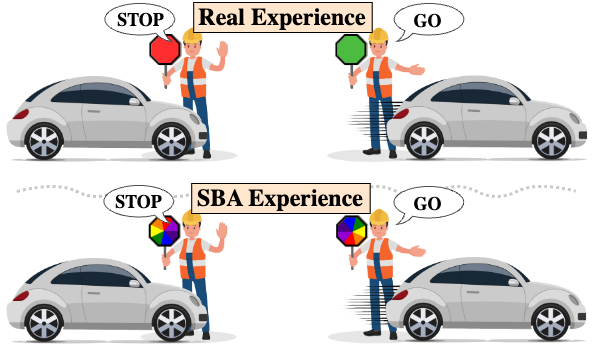}
  \caption{Augmenting conventions of other agents. The driver stops at \textcolor{dark_red}{red} and drives on \textcolor{dark_green}{green} (top), but with SBA, our agent \textit{sees} the driver stopping and starting with many \textcolor{dark_red}{c}\textcolor{dark_orange}{o}\textcolor{dark_yellow}{l}\textcolor{dark_green}{o}\textcolor{dark_blue}{u}\textcolor{dark_purple}{r}\textcolor{dark_pink}{s} (bottom).}
  \label{fig:traffic_conductor}
  \vspace{-15pt}
\end{wrapfigure}

\section{Introduction}

Humans and AI agents alike employ a diverse range of \textit{conventions} when interacting with one another. These conventions facilitate communication and coordination, which are crucial for effective teamwork in many multi-agent settings. They range in complexity from knowing which side of the road to drive on to coordinating using a shared language. For agents to effectively coordinate with others—particularly in contexts where strategic conventions vary—they must develop an understanding of these conventions, especially when coordination failures can lead to severe consequences such as vehicle collisions.

The challenge of aligning to previously unseen teammates has been formalised as ad hoc teamwork (AHT)~\citep{stone2010ad}. One method of training and evaluating an AHT agent is to use reinforcement learning (RL) to learn a \textit{best response} (BR) to a training population of teammates \citep{fudenberg1991game} (more formally defined in Section~\ref{sec:aht_objective}) and then evaluate against a test set of held-out agents. The key challenge, therefore, is \textit{generalising} to unseen policies after only being exposed to a subset of possible strategies during training, a problem that reflects the practical difficulties of coordinating effectively in diverse human-AI scenarios.

Furthermore, due to symmetries, the space of possible conventions is often combinatorial even in simple environments, making it computationally challenging to compute the \textit{best response} even if the training population were sufficiently large to cover the distribution. Children reduce this computational burden by transferring existing knowledge to equivalent symmetries through higher-level reasoning~\citep{beasty1987role}. In contrast, chimpanzees have been observed to fail these symmetry tests~\citep{dugdale2000testing}. AI agents also struggle with this, from computer vision models that fail to generalise to different coloured images~\citep{galstyan2022failure} to cooperative agents that cannot recognise when teammates are using symmetry-equivalent conventions~\citep{hu2020other}.

To address this within a human-AI alignment framework, we introduce \textit{symmetry-breaking augmentations} (SBA), a policy augmentation technique that alters the behaviour of agents in the training pool by making them break symmetries in various ways. SBA acts as an operator that can be applied to other agents in an environment, flipping their behaviours along environmental symmetries. When symmetries are present, this combinatorially amplifies the range of conventions to which the ad hoc agent is exposed. Thus, even with a relatively small training population, SBA enables RL agents to learn to adapt to a much more diverse set of conventions during training. This technique not only improves test-time performance but also ensures the agent acts predictably in relation to environmental symmetries, making it easier for humans to adapt to the its behaviour.

For example, consider a traffic conductor learning to direct drivers. Initially, the conductor is unaware of the colour conventions drivers may use for stop and go. As shown in Figure~\ref{fig:traffic_conductor}, SBA can be used to create new training experiences by altering the observed colours. This prevents the conductor from overfitting to the potentially limited conventions of its training partners and provides experiences that enable adaptation to different conventions in the future.

SBA is closely related to zero-shot coordination (ZSC) approaches such as Other Play \citep{hu2020other} and Equivariant Networks \citep{muglich2022equivariant}. However, since these approaches aim to learn policies that are invariant to environmental symmetries, they are not directly applicable to the AHT setting, where an agent must be able to coordinate with teammates that \textit{do} use symmetry-breaking conventions. SBA also differs from population-based ZSC and AHT approaches \citep{lupu2021trajectory, rahman2023minimum} in that it aims to generalise from a provided set of partners rather than generating a sufficiently diverse set of partners from scratch. SBA could be applied in conjunction with these population-based approaches to further increase the diversity of policies.

Since, SBA is most effective when the agents in the training population use symmetry-breaking conventions, we introduce the \textit{Augmentation Impact} (AugImp) metric, which measures the extent to which a specific augmentation alters policies within a population. This enables us to analyse a population prior to training to predict how effective SBA will be at improving AHT performance.

We demonstrate how SBA improves performance in both a simple matrix game and the card game Hanabi. In Hanabi, independently trained agents frequently develop incompatible conventions, even when trained using the same algorithm~\citep{hu2020other}. Since Humans also use a diverse range of conventions when playing Hanabi, a successful AHT Hanabi agent needs to quickly infer and adapt to the conventions of its teammates while avoiding triggering unexpected responses. We experiment by training an AHT agent with existing Hanabi populations of simplified action decoder (SAD)~\citep{hu2019simplified} and independent Q-learning (IQL)~\citep{tan1993multi} policies. We show that in Hanabi, SBA leads to improvements of up to 17\% in game score when adapting to previously unseen teammates from the same population. Additionally, we show that SBA improves performance when generalising outside of the training distribution to populations of Other Play (OP)~\citep{hu2020other} and Off-Belief Learning (OBL) agents~\citep{hu2021off}.

To summarise, our contributions are:
\begin{itemize}[leftmargin=*, noitemsep, topsep=0pt]
    \item SBA, a general method of augmenting a training population for AHT that amplifies the diversity of conventions the agent is exposed to during training.
    \item A general metric that measures the effectiveness of policy augmentation techniques for AHT by assessing how much they diversify behaviours in a training population.
    \item Evaluation of SBA in Hanabi, demonstrating state-of-the-art performance for AHT.
\end{itemize}

\section{Background}

\subsection{Dec-POMDPs}

We formalise the cooperative multi-agent setting as a decentralised partially-observable Markov Decision Process (Dec-POMDP)~\citep{nair2003taming}. 
The Dec-POMDP, $G$, is a 9-tuple $(\mathcal{N}, \mathcal{S}, \{\mathcal{A}^i\}_{i=1}^n, \{\mathcal{O}^i\}_{i=1}^n, \mathcal{T}, \mathcal{U}, \mathcal{R}, T, \gamma)$, with finite sets $\mathcal{N}, \mathcal{S}, \{\mathcal{A}^i\}_{i=1}^n, \{\mathcal{O}^i\}_{i=1}^n$, respectively denoting the set of agents, states, actions, and observations, where $i$ denotes the set pertaining to agent $i \in \mathcal{N} = \{1, \dots, n\}$. 
$\mathcal{A}^i$ and $\mathcal{O}^i$ are the set of actions and observations for agent $i$, and $a^i \in \mathcal{A}^i$ and $o^i \in \mathcal{O}^i$ are a specific action and observation that agent $i$ may take and observe.
We also write $\mathcal{A} = \times_{i=1}^nA^i$ and $\mathcal{O} = \times_{i=1}^nO^i$, as the sets of joint actions and observations, respectively. 
$s_t \in \mathcal{S}$ is the state at time $t$, and $a_t \in \mathcal{A}$ is the joint action of all agents at time $t$, which changes the state according to the transition distribution $s_{t+1} \sim \mathcal{T}(\cdot|s_t, a_t)$. 
The subsequent joint observation of the agents, $o_{t+1} \in \mathcal{O}$, is distributed according to $o_{t+1} \sim \mathcal{U}(\cdot|s_{t+1}, a_t)$, where $\mathcal{U} = \times_{i=1}^n\mathcal{U}^i$. 
At time $t$, the joint observation $o_t$ is appended to the trajectory $\tau_t=(o_1, a_1, \dots, o_{t-1}, a_{t-1}, o_t)$, and each agent $i$ individually decides its own action $a_t^i$ based on its policy $\pi^i(a_t^i|\tau_t^j)$, which is conditioned on its action-observation history (AOH) $\tau_t^i=(o_1^i, a_1^i, \dots, o_{t-1}^i a_{t-1}^i, o_t^i)$. 
$\pi^i$ represents agent $i$'s component of the decentralized joint policy $\pi \in \Pi$, where $\Pi$ is the set of all possible joint-policies in the environment $G$. 
When $G$ transitions to state $s_{t+1}$, all agents receive a common reward $r_{t+1} \in \mathbb{R}$ according to the distribution $r_{t+1} \sim \mathcal{R}(\cdot|s_{t+1}, a_t)$. 
The behaviour of a joint-policy $\pi$ is characterised by the distribution of trajectories $\tau$ it produces, and, taking into account the time horizon $T$ and discount factor $\gamma \in [0,1]$, is optimal if it maximises the expected return:
\begin{equation} \label{eqn:expected_return}
J(\pi) = \mathbb{E}_{\tau \sim \pi}[\sum_{t=1}^T{\gamma^{t-1}r_t}].
\end{equation}

\subsection{Ad Hoc Teamwork} \label{sec:aht_objective}

Ad hoc teamwork (AHT) is the problem of creating an agent that is able to collaborate effectively with a group of novel teammates. This has been a long-standing challenge in the field of artificial intelligence~\citep{stone2010ad, bard2020hanabi}.

We use $\pi_A$ to represent the AHT agent, and $\pi_j$ for the teammate's joint-policy. To evaluate the performance of our AHT joint-policy $\pi_A = (\pi_A^1, \dots, \pi_A^n)$ and the teammate joint-policy $\pi_j$, using Equation~\ref{eqn:expected_return}, we obtain the average expected AHT return by matching each individual component of $\pi_A$, i.e. $\pi^i_A$, with all other $n - 1$ components of $\pi_j$,  i.e. $\pi^i_j$. 
This objective is formalised by:
\begin{equation} \label{eqn:aht_expected_return}
    J_{AHT}(\pi_A,\pi_j) = \frac{1}{n} \big( J(\pi^1_A,\pi^2_j, \dots, \pi^n_j) + \dots + J(\pi^1_j, \dots, \pi^{n-1}_j, \pi^n_A) \big),
\end{equation}
Our AHT agent $\pi_A$ learns a best-response $\pi_A^*$ to a training set $\Pi^{train} \in \Pi$ by interacting with each policy $\pi_j \in \Pi^{train}$, and maximising Equation~\ref{eqn:aht_expected_return}. We formally define this objective as:
\begin{equation} \label{eqn:aht_train_definition}
    \pi_A^*(\Pi^{train}) =
    \underset{\pi_A}{\text{argmax}}~\mathbb{E}_{\pi_j \sim \Pi^{train}}[J_{AHT}(\pi_A, \pi_j)],
\end{equation}
where $\pi_j$ is sampled uniformly from $\Pi^{train}$. 
The learned AHT policy, $\pi_A^*(\Pi^{train})$, is then evaluated using the robustness measure, $M_{\Pi^{eval}}(\pi_A^*(\Pi^{train}))$, which evaluates Equation~\ref{eqn:aht_expected_return} while interacting with previously unseen policies from the evaluation policy set $\Pi^{eval} \in \Pi$. This measure is formally given by:
\begin{equation} \label{eqn:aht_eval_definition}
    M_{\Pi^{eval}}(\pi_A^*(\Pi^{train})) = \\
    \mathbb{E}_{\pi_j \sim \Pi^{eval}}J_{AHT}(\pi_A^*(\Pi^{train}), \pi_j),
\end{equation}
where $\pi_j$ is sampled uniformly from $\Pi^{eval}$.

\subsection{Equivalence Mappings} \label{sec:equiv_mappings}

To improve coordination with unseen teammates in the Dec-POMDP setting, domain knowledge can be exploited to increase the variety of conventions present in the training set.
To achieve this we use a class of \textit{equivalence mappings} (symmetries)~\citep{hu2020other}, $\Phi$, for a given Dec-POMDP $G$, such that each $\phi \in \Phi$ is an automorphism of $\mathcal{S}$, $\mathcal{A}$, and $\mathcal{O}$ onto itself, and leaves $G$ unchanged up to relabeling such that the environment dynamics and rewards function stay the same:
\begin{align} \label{eqn:symmetry_operator_full}
    \phi \in \Phi \Longleftrightarrow \mathcal{T}(\phi(s_{t+1}) | \phi(s_t), \phi(a_t)) &= \mathcal{T}(s_{t+1} | s_t, a_t) \nonumber \\
    \wedge \mathcal{R}(\phi(r_{t+1}) | \phi(s_{t+1}), \phi(a_t)) &= \mathcal{R}(r_{t+1} | s_{t+1}, a_t) \nonumber \\
    \wedge \mathcal{U}^i(\phi(o^i_{t+1}) | \phi(s_{t+1}), \phi(a_t)) &= \mathcal{U}^i(o^i_{t+1} | s_{t+1}, a_t) \nonumber \\
    \text{where equalities apply } \forall s_{t+1}, s_t \in \mathcal{S},& a_t \in \mathcal{A}, i \in \mathcal{N}.
\end{align}
For ease of notation, $\phi$ is shorthand for 
\begin{equation} \label{eqn:symmetry_operator}
    \phi \in \Phi = \{\phi_{\mathcal{S}}, \phi_{\mathcal{A}}, \phi_{\mathcal{O}}\}, 
\end{equation}
where each $\phi \in \Phi$ acts on trajectories as
\begin{equation} \label{eqn:symmetry_operator_aoh}
    \phi(\tau_t) = (\phi(o_0), \phi(a_0), \dots, \phi(a_{t - 1}), \phi(o_t)),
\end{equation}
and acts on policies as
\begin{equation} \label{eqn:symmetry_operator_actions}
    \hat{\pi} = \phi(\pi) \Longleftrightarrow \hat{\pi}(\phi(a) | \phi(\tau)) = \pi(a | \tau).
\end{equation}

\begin{wrapfigure}{r}{0.5\textwidth}
    \centering
    \vspace{-10pt}
    \includegraphics[width=0.30\textwidth]{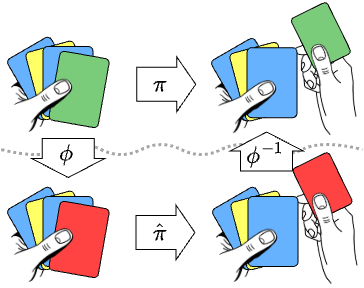}
    \caption{The $\phi$ operator converts \textcolor{dark_green}{green} observations to \textcolor{dark_red}{red} (left), and $\phi^{-1}$ inversely converts \textcolor{dark_red}{red} actions back to \textcolor{dark_green}{green} (right). In this game \textcolor{dark_red}{red} and \textcolor{dark_green}{green} are symmetrically-equivalent, so the application of $\phi$ and $\phi^{-1}$ leaves the game unchanged up to relabelling.}
    \label{fig:augmented_card_colours}
    \vspace{-15pt}
\end{wrapfigure}

Policies $\pi$, $\hat{\pi}$ in Equation~\ref{eqn:symmetry_operator_actions} are said to be symmetry-equivalent to one another with respect to $\phi$.
For every symmetry operator $\phi$ in the automorphism group $\Phi$, there exists an inverse operator $\phi^{-1} \in \Phi$ such that $\phi \circ \phi^{-1} = \phi^{-1} \circ \phi = e$, where $e$ is the identity automorphism of $\Phi$, and $\circ$ denotes function composition. Illustrated in Figure~\ref{fig:augmented_card_colours}, using $\phi$, the augmented policy $\hat{\pi}$ experiences a symmetrically-equivalent version of $\tau_t$, and its actions are converted back to their original mapping with $\phi^{-1}$. 
While OP uses $\phi$ to prevent symmetry-breaking conventions for ZSC, our work applies $\phi$ to the AHT setting, improving robustness (Equation~\ref{eqn:aht_eval_definition}) by increasing the variety of conventions present in the training set $\Pi^{train}$.

\section{Symmetry-Breaking Augmentations}

One of the biggest challenges in AHT is predicting the conventions that test-time policies will use and determining how a training population should be selected so that its best response is robust to these conventions. In the following, we introduce SBA, a method that addresses this problem by augmenting the training population through the random matching of the AHT agent with symmetry-equivalent policies of training teammates. We will discuss SBA both as a formal method and as a scalable algorithm-agnostic framework applicable to the deep RL setting.

\subsection{SBA Objective}

We start by introducing the SBA learning rule which uses the set of equivalence mappings $\phi \in \Phi$ from Section~\ref{sec:equiv_mappings} to diversify an existing training population $\Pi^{train}$ such that the best-response is robust to a larger variety of conventions.

The intuitive approach is to apply a different $\phi$ to each teammate $\pi^i_j$ to maximise the variety of teams encountered.
However, in a fully-collaborative Dec-POMDP, using different $\phi$'s will break the coordination between each $n - 1$ components, $\pi^i_j$, so the same $\phi$ needs to be applied instead.
Moreover, if $\pi^i_j$ is a physical agent acting in the real world, then $\phi$ can't easily be applied as it requires us to modify its actions and observations.

\textbf{Lemma 1.} $J(\pi) = J(\phi(\pi)) \forall \phi \in \Phi, \pi \in \Pi$ 

This Lemma shows that the expected return of a joint-policy $\pi$ is equal to the expected return when $\phi$ is
applied $\pi$.

Proof in Appendix~\ref{apx:proofs_lemma1}.

\textbf{Proposition 1.}
The expected AHT return when $\phi$ is applied to $\pi$ is equal to the expected AHT return when the inverse $\phi^{-1}$ is applied to each of the $\pi^i_j$ teammate policies.

Proof in Appendix~\ref{apx:proofs_prop1}.

By applying $\phi$ to $\pi_A$, we're guaranteed to always be able to alter $\pi_j$'s perceived conventions, and its application is of order $O(1)$. 
With this, our AHT agent $\pi_A$ learns the optimal best-response ($SBA^*$) to $\Pi^{train}$ that has been augmented with $\Phi$, by interacting with each policy $\pi_j \in \Pi^{train}$, applying $\phi \in \Phi$ to $\pi_A$, and maximising the expected AHT return~(Equation~\ref{eqn:aht_expected_return}). We formally define this objective as:
\begin{equation} \label{eqn:sba_definition}
    \pi_A^*(\Pi^{train}) =     \underset{\pi}{\text{argmax}}~\mathbb{E}_{\pi_j \sim \Pi^{train}, \phi \sim \Phi}[J_{AHT}(\phi(\pi_A), \pi_j)],
\end{equation}
where $\pi_j$ and $\phi$ are uniformly sampled from $\Pi^{train}$ and $\Phi$ respectively. To evaluate the robustness when interacting with an unseen evaluation set $\Pi^{eval}$, we use the same robustness measure from Equation~\ref{eqn:aht_eval_definition}, $M_{\Pi^{eval}}(\pi_A^*(\Pi^{train}))$, and also apply equivalence mappings to the evaluation policies to reduce the evaluation variance (effectively generating a larger test-population). This measure is formally given by:
\begin{align} \label{eqn:aht_eval_sba_definition}
    M_{\Pi^{eval}}(\pi_A^*(\Pi^{train})) = \mathbb{E}_{\pi_j \sim \Pi^{eval}, \phi \sim \Phi}J_{AHT}(\phi(\pi_A^*(\Pi^{train})), \pi_j),
\end{align}
where $\pi_j$ and $\phi$ are sampled uniformly from $\Pi^{eval}$ and $\Phi$.

\subsection{Algorithm}

\begin{wrapfigure}{r}{0.5\textwidth}
\vspace{-10pt}
\centering
\resizebox{0.5\textwidth}{!}{
\begin{tikzpicture} [
    circleStyle/.style={circle, draw, inner sep=0, minimum size=10mm},
    rectangleStyle/.style={rounded corners, draw, inner sep=0, minimum width=10mm, minimum height=8mm},
    arrowStyle/.style={-{>[length=3mm, width=2mm]}}
]
    \node                 (st-1) {};
    \node[circleStyle]    (st)  [right=of st-1]  {\Large $s_t$};
    \node[circleStyle]    (st+1) [right=of st]  {\Large $s_{t+1}$};
    \node                 (st+i) [right=of st+1]  {\Large $\dots$};
    \node[circleStyle]    (st+n-1) [right=of st+i]  {\large $s_{t+n-1}$};
    \node[circleStyle]    (st+n) [right=of st+n-1] {\Large $s_{t+n}$};
    \node                 (st+n+1) [right=of st+n] {};
    \node[rectangleStyle] (pi)  [below=of st]  {\Large $\hat{\pi}^i_A$};
    \node[rectangleStyle] (pi-i+1) [above=of st+1] {\Large $\pi^{i+1}_j$};
    \node[rectangleStyle] (pi-i+n-1) [above=of st+n-1] {\large $\pi^{i+n-1}_j$};

    \draw[arrowStyle] (st-1.east) to (st.west);
    \draw[arrowStyle] (st.east) to node[below] {\large $r_t$} (st+1.west);
    \draw[arrowStyle] (st+1.east) to node[below] {\large $r_{t+1}$} (st+i.west);
    \draw[arrowStyle] (st+i.east) to node[below] {\large $r_{t+n-1}$} (st+n-1.west);
    \draw[arrowStyle] (st+n-1.east) to node[below] {\large $r_{t+n}$} (st+n.west);
    \draw[arrowStyle] (st+n.east) to (st+n+1.west);
    \draw[arrowStyle] ([xshift=-2mm]st.south) to node[left] {\large\textcolor{medium_red}{$\phi(\tau_t^i)$}} ([xshift=-2mm]pi.north);
    \draw[arrowStyle] ([xshift=2mm]pi.north) to node[right] {\large\textcolor{medium_red}{$\phi^{-1}(a_t^i)$}} ([xshift=2mm]st.south);
    \draw[arrowStyle] ([xshift=-2mm]st+1.north) to node[left] {\large $\tau_{t+1}^{i+1}$} ([xshift=-2mm]pi-i+1.south);
    \draw[arrowStyle] ([xshift=2mm]pi-i+1.south) to node[right] {\large $a_{t+1}^{i+1}$} ([xshift=2mm]st+1.north);
    \draw[arrowStyle] ([xshift=-2mm]st+n-1.north) to node[left] {\large $\tau_{t+n-1}^{i+n-1}$} ([xshift=-2mm]pi-i+n-1.south);
    \draw[arrowStyle] ([xshift=2mm]pi-i+n-1.south) to node[right] {\large $a_{t+n-1}^{i+n-1}$} ([xshift=2mm]st+n-1.north);
    \draw[arrowStyle, dashed] (st+n.south) .. controls +(down:16mm) and +(right:8mm) .. node[below, xshift=6mm] {\Large $r_t + \dots + r_{t+n}$, \textcolor{medium_red}{$\phi(\tau_{t+n}^i)$}} (pi.east);
\end{tikzpicture}}
\caption{\textit{Symmetry-breaking augmentations} for an $n$-player Dec-POMDP. The equivalence map \textcolor{dark_red}{$\phi$} is only applied to the observations and actions of our AHT agent $\pi_A$, not the teammate policy $\pi_j$.}
\label{fig:state_diagram}
\vspace{-10pt}
\end{wrapfigure}
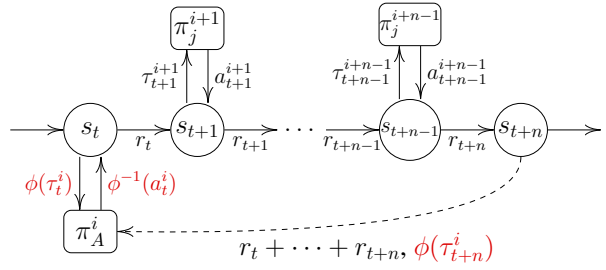

The idea behind SBA is simple: As shown in Figure~\ref{fig:state_diagram}, each of the AHT agents observations $o^i_t$ are mapped with $\phi$ to an equivalent state with relabelled symmetries, $\pi^i_A$ chooses an action $a^i_t$, and the action is inversely relabelled with $\phi^{-1}$ before being applied to the environment. The permuted observations and actions are appended to $\pi^i_A$'s AOH $\tau^i$, which is used to update the model. Notice that $\phi$ influences the agent's actions and observations, not the environment dynamics, and therefore any standard RL learning algorithm can be used to update the model, like DQN~\citep{mnih2015human}, DDPG~\citep{lillicrap2015continuous}, A3C~\citep{mnih2016asynchronous}, or PPO~\citep{schulman2017proximal}.

\begin{wrapfigure}{r}{0.56\textwidth}
\vspace{-20pt}
\begin{algorithm}[H]
\begin{minipage}{\textwidth}
    \caption{Symmetry-Breaking Augmentations}
    \SetAlgoLined
    \DontPrintSemicolon
    \SetKwInOut{Input}{Input}
    \SetKwInOut{Initialise}{Initialise}
    \textbf{Input:} algorithm $\mathbb{A}$, Dec-POMDP $G$, population $\Pi^{train}$\;
    \textbf{Initialise:} $\mathbb{A}$, equivalence mappings $\Phi$ from $G$\;
    \For{each episode} {
        $\pi_{-j} \gets$ teammate policy sampled from $\Pi^{train}$\;
        $\phi \gets$ equivalence mapping sampled from $\Phi$\;
        $s_0, \tau_0 \gets$ initial state and history\;
        \BlankLine
        \For{each step $t$} {
            append observation $o_t^i$ from $s_t$ to AOH $\tau_t^i$\;
            $a_t^i \gets$ sample action using $\mathbb{A}$: $\phi^{-1}(\pi_j^i(\cdot|\phi(\tau_t^i)))$ \;
            append action $a_t^i$ to AOH $\tau_t^i$\;
            \For{each teammate component $\pi_{-j}^{-i}$} {
                append observation $o_{t}^{-i}$ from $s_t$ to AOH $\tau_t^{-i}$\;
                $a_t^{-i} \gets$ sample action from $\pi_{-j}^{-i}(\cdot|\tau_t^{-i})$\;
                append action $a_t^{-i}$ to AOH $\tau_t^{-i}$\;
            }
            take joint-action $a_t$, observe $r_t$, and $s_{t+1}$\;
        }
        \For{each AHT agent turn $t$} {
            $r_{t:t+n} \gets$ sum rewards from $r_t$ to $r_{t + n}$\;
            $\mathbb{T} \gets$ transition $(o_t^i, a_t^i, r_{t:t+n}, o_{t+n}^i)$\;
            perform one step of optimisation using $\mathbb{A}$ and $\mathbb{T}$\;        
        }
    }
    \label{alg:symmetry_breaking_augmentations}
\end{minipage}
\end{algorithm}
\vspace{-65pt}
\end{wrapfigure}

In the simplest version of our algorithm, all of the AHT agents' transitions, $\mathbb{T} = (\tau_t^i, a_t^i, r_{t:t+n}, \tau_{t+n}^i)$ are stored, where $r_{t:t+n} = \sum_{t' = t}^{t+n}r_{t'}$ is the sum of all rewards received between time step $t$ and $t+n$, and $\mathbb{T}$ is used to update the model. See Algorithm~\ref{alg:symmetry_breaking_augmentations} for a more formal description.

\subsection{Augmentation Impact} \label{sec:augmentation_impact}

We aim to increase the diversity of training partners by applying SBA to each member of the training population, and hypothesize that this will lead to better generalisation in the AHT setting. However, if the training agents barely rely on symmetry-based conventions, SBA will have little effect, i.e. $\phi(\pi_j) \approx \pi_j, \forall \pi_j \in \Pi^{train}, \forall \phi \in \Phi$.

Since training an AHT agent can be expensive, it is useful to quantify how much SBA will diversify a population prior to training.
For this, we introduce \textit{Augmentation Impact} (AugImp), a metric that takes a population $\Pi$, a set of equivalence mappings $\Phi$, and for each pair of policies $\pi_1,\pi_2 \in \Pi$, measures the expected absolute difference of the crossplay scores~\citep{lupu2021trajectory} with and without each augmentation $\phi \in \Phi$ being applied to one of the policies, $\pi_1$. AugImp is formalised by: 
\begin{equation} \label{eqn:augmentation_impact}
    \text{AugImp}(\Pi, \Phi) = \mathbb{E}_{\pi_1 \sim \Pi, \pi_2 \sim \Pi, \phi \sim \Phi} [ \nonumber |J_{\text{XP}}(\phi(\pi_1), \pi_2) - J_{\text{XP}}(\pi_1, \pi_2)|] 
\end{equation}
where $\pi_1, \pi_2$, and $\phi$ are uniformly sampled from $\Pi$ and $\phi$ respectively. The bigger the AugImp score, the more $\Pi$ is diversified by $\Phi$, and the better effect that SBA will have when training an AHT agent with that population.

\section{Iterated Lever Game Experiments}

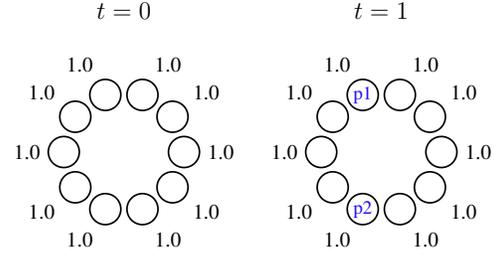
\begin{wrapfigure}{r}{0.5\textwidth}
    \centering
    \vspace{-20pt}
\resizebox{0.5\textwidth}{!}{%
\begin{tikzpicture}[align=center]
    \node [circle, minimum size=2cm] (c) {};
    \node [circle, minimum size=2cm] (c2)  [right=2.3cm of c] {};
    \node (cname) [above=1.1cm of c] {\large $t=0$};
    \node (c2name) [above=1.1cm of c2] {\large $t=1$};
    
    \foreach \circle in {c,c2}{
        \foreach \a in {1,...,10}{
            \newcommand{\nodetext}{}
            \ifthenelse{\equal{\circle}{c2}}
                {\ifthenelse{\equal{\a}{3}}
                    {\renewcommand{\nodetext}{\small\textcolor{blue}{p1}}}
                    {\ifthenelse{\equal{\a}{7}}
                        {\renewcommand{\nodetext}{\small\textcolor{blue}{p2}}}{}
                    }
                }{}
            \node[
                draw, thick, 
                inner sep=0,
                minimum size=5.2mm, 
                circle,
                label={\a*360/10:1.0} 
                ] at (\circle.\a*360/10)
                {\nodetext};
        }
    }
\end{tikzpicture}
}
\caption{In the \textit{iterated lever coordination game} agents can see 
    what \textcolor{blue}{actions} were previously taken. The game highlights 
    the difficulty of adapting to conventions not seen during training.}
\label{fig:lever_game}
\vspace{-10pt}
\end{wrapfigure}

We first test SBA in a simple fully cooperative environment where agents are tasked to coordinate by pulling one of ten possible levers. As shown in Figure~\ref{fig:lever_game}, a reward of $1$ is paid out if both players pick the same lever, otherwise they get nothing. The game is played twice, but in the second round the players are able to see what lever their partner previously pulled. If agents could coordinate beforehand they would always pull the same lever, but when playing with an unknown teammate there is no way to coordinate on the first round. To apply SBA to the lever game, since a permutation of the levers leaves the game unchanged, we use this as our class of symmetries.

\begin{wrapfigure}{r}{0.5\textwidth}
    \centering
    \vspace{-20pt}
    \includegraphics[width=0.5\textwidth]{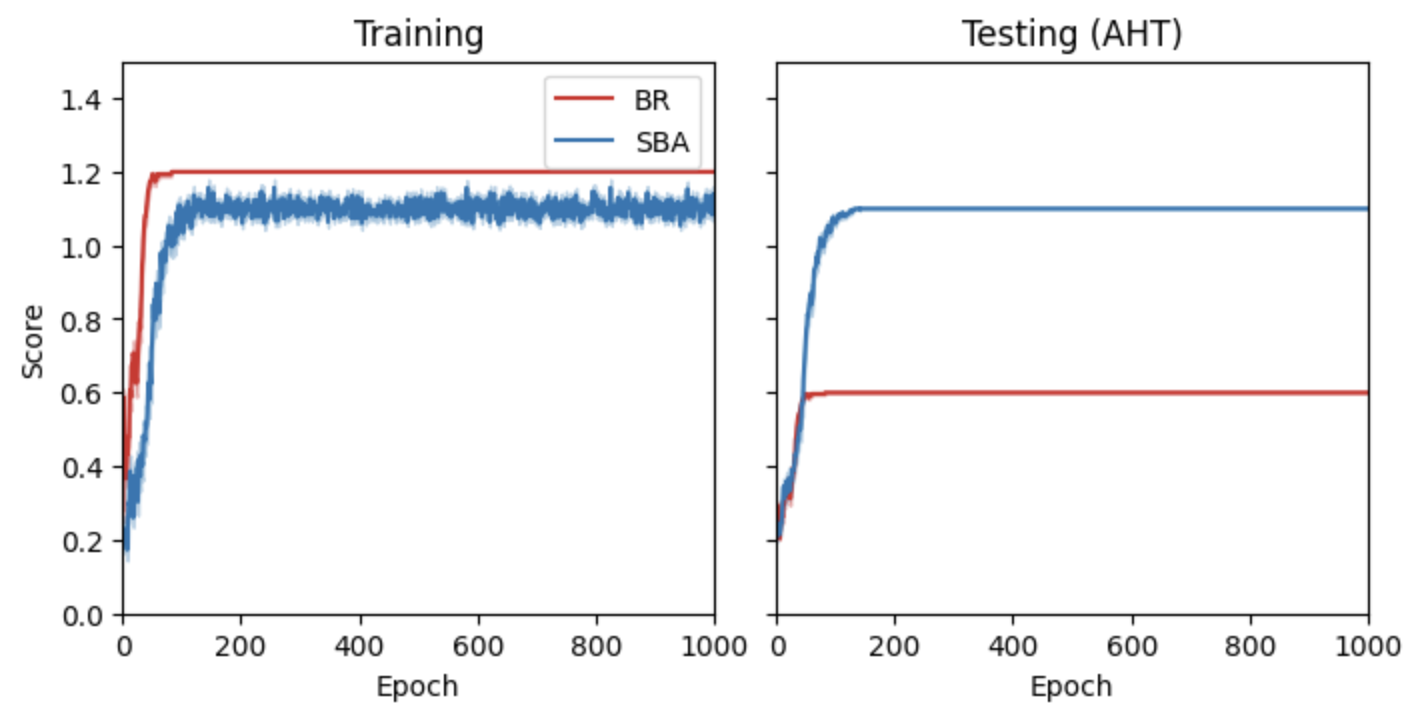}
    \caption{Training curves for the \textit{iterated lever coordination game}. 
    Shown is the mean, shading is the standard error of the mean, across 
    $30$ different seeds. SBA improves test performance because it exposes 
    the agent to more conventions during training.}
    \label{fig:lever_game_results}
\end{wrapfigure}

We train our AHT agent with a population of five different teammates that each deterministically pull one of the levers, and evaluate with ten policies that pull all ten levers. We refer to Appendix~\ref{apx:lever_experiment_setup} for more details on the implementation. The code is available online without downloading: \url{https://bit.ly/lever-game-sba}.

The results are shown in Figure~\ref{fig:lever_game_results}, where, as expected, all agents randomly choose a lever in the first round. In the second round, during training the baseline (BR) always successfully switches to the correct lever for a return of $1.2$, but only scores $0.6$ at test time because it doesn't expect the other five levers to be pulled. Our SBA agent, however, experiences all levers during training, so is able to adapt to all teammates for a total score of $1.1$ in both training and testing. 

\section{Hanabi Experiments}

\begin{wrapfigure}{r}{0.5\textwidth} 
    \centering
    \vspace{-50pt}
    \includegraphics[width=0.4\textwidth]{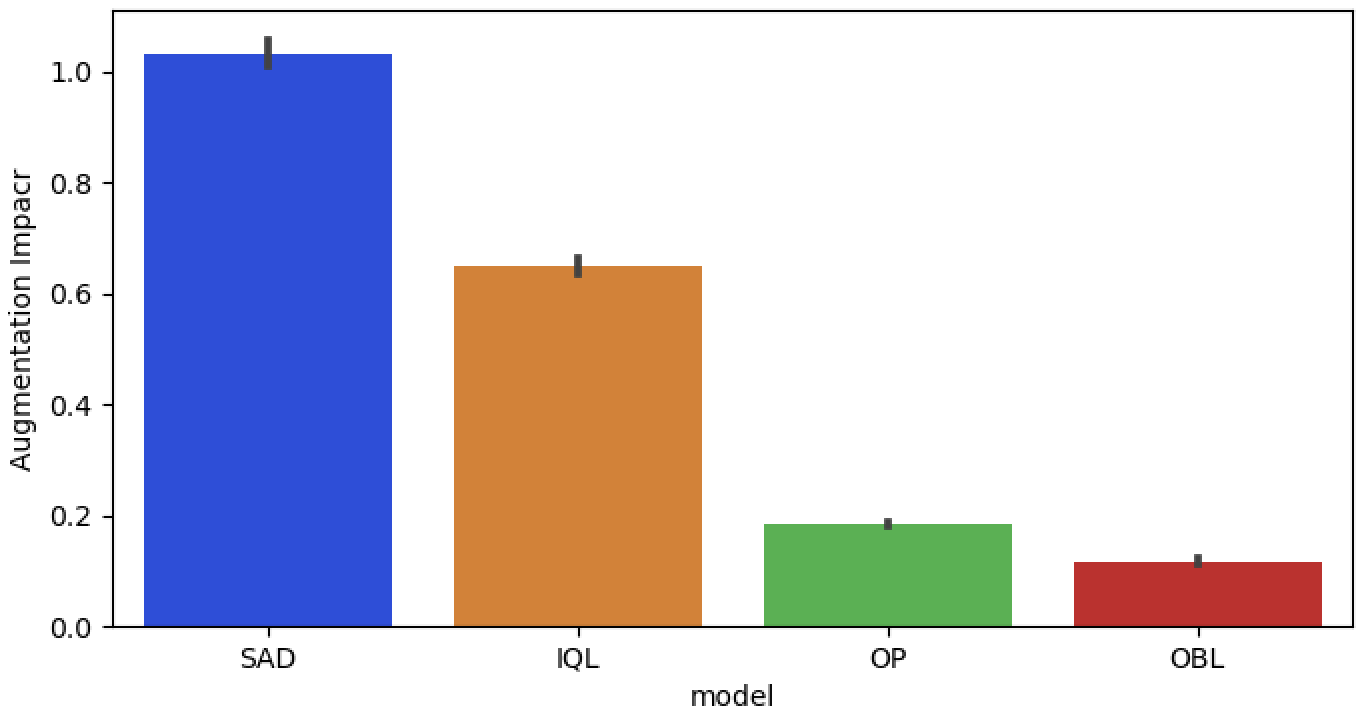}
    \caption{\textit{Augmentation Impact} (AugImp) for Hanabi populations. SAD and IQL populations have a larger AugImp than OP and OBL, because they contain policies with more symmetry-breaking conventions.}
    \label{fig:augmentation_impact}
    \vspace{-20pt}
\end{wrapfigure}

We now test SBA in Hanabi. Hanabi is a fully-cooperative, partially-observable card game~\citep{bard2020hanabi} for MARL, theory of mind, and AHT research. In Hanabi, players cannot see their own cards and must rely on limited clues from teammates to cooperatively build five coloured decks in ascending order without triggering bombs. For the full rules of the game, please see Appendix~\ref{apx:hanabi_rules}. For a full description of our experiment setup, please see Appendix~\ref{apx:hanabi_experiment_setup}.

\subsection{Teammate Selection} \label{sec:teammate_selection}

Before training our agent, a teammate policy population needs to be selected. Since there exists a range of pre-trained Hanabi agent populations online, we use AugImp (Equation~\ref{eqn:augmentation_impact}) to guide the selection. We analyse the AugImp scores for four populations: 13 simplified action decoder (SAD), 12 independent Q-learning (IQL), 12 other-play (OP) models, and 5 off-belief learning (OBL) models (pre-trained weights for these models are available on GitHub\footnote{\href{https://github.com/facebookresearch/hanabi_SAD}{https://github.com/facebookresearch/hanabi\_SAD}}\footnote{\label{fnt:obl}\href{https://github.com/facebookresearch/off-belief-learning}{https://github.com/facebookresearch/off-belief-learning}}).

To calculate the crossplay scores for each pair of policies $\pi_1$ and $\pi_2$ and each augmentation $\phi$, we take the mean score over 1000 games. Figure~\ref{fig:augmentation_impact} shows the AugImp scores for each population. The scores for SAD and IQL are much larger than OP and OBL, which is expected because OP and OBL are designed to use conventions that don't break symmetries. For a complete breakdown of the AugImp score distributions for each policy pairing, see Appendix~\ref{apx:augmentation_impact}. Since SBA increases the diversity of SAD and IQL, we use these populations for training because we expect the largest AHT performance improvement.

\subsection{Best response agents}
We train AHT agents using the pre-trained SAD and IQL populations, and evaluate their ad hoc generalisation to held-out partners. We create a number of testing and training splits for each population. Each population is randomly divided into \textit{small}, \textit{medium}, or \textit{large} training/test splits: \textit{small} splits have 1 training policy, \textit{medium} splits have 6, and \textit{large} training sets contain all but 2 policies from the population. The remaining policies form the test set for that split. Smaller split sizes are more challenging, as our AHT agents are exposed to a more limited set of partners during training. We randomly sample 10 different partitions for \textit{medium} and \textit{large} training sets, and run all possible partitions for \textit{small} (13 for SAD and 12 for IQL). See Appendix~\ref{apx:population_splits} for details on train-test splits.

We train AHT agents using SBA on pre-existing populations rather than generating our population from scratch as in \citet{lupu2021trajectory, rahman2023minimum}. While this approach is limited in that it requires pre-existing training policies to be available, it has the advantage of being a natural way for us to specify a prior over strategies that we want our AHT agent to specialise in. This is important in the context of Hanabi, where the space of possible strategies is large. We also believe that this approach is sufficient to demonstrate the effectiveness of SBA at allowing AHT agents to generalise to symmetry-equivalent held-out partners.

\subsection{Ad Hoc Teamwork Results}

\begin{wrapfigure}{r}{0.5\textwidth} 
\vspace{-35pt}
\centering
\begin{adjustbox}{width=0.50\textwidth}
\begin{tabular}{llll}
\toprule
Train Size & Agent & SAD $\uparrow$ & IQL $\uparrow$ \\
\midrule
\multirow{2}{*}{small} & BR         & 8.15 $\pm$ 1.28     & 11.52 $\pm$ 1.08 \\
                      & SBA (ours) & \bfres{9.12}{1.42}   & 11.84 $\pm$ 0.99 \\
\midrule
\multirow{3}{*}{medium}  & Gen. Belief & 12.47 $\pm$ 1.02 & - \\
                      & BR         & 13.09 $\pm$ 0.49    & 15.04 $\pm$ 0.37 \\
                      & SBA (ours) & \bfres{15.40}{0.49}   & \bfres{16.08}{0.42} \\          
\midrule
\multirow{2}{*}{large} & BR         & 14.69 $\pm$ 1.05     & 15.34 $\pm$ 0.80 \\
                      & SBA (ours) & \bfres{16.34}{1.29} & 15.95 $\pm$ 0.71 \\
\midrule
& OP         & 3.26 $\pm$ 1.20     & 12.00 $\pm$ 0.51 \\
\bottomrule
\end{tabular}
\end{adjustbox}
\caption{SBA performance in Hanabi. The reported score for generalized 
beliefs (Gen. Belief) on the SAD \textit{medium} split size. Shown is the 
standard error of the mean (s.e.m) across the \textit{small}, \textit{medium}, 
and \textit{large} training train-test splits.}
\label{table:sba_aht_results}
\vspace{-15pt}
\end{wrapfigure}

Here we examine the impact of SBA on AHT performance to the held-out test agents for SAD and IQL. For each of the train/test splits outlined above, we train a standard best response with and without SBA. Each agent is evaluated on the held out test agents from its training split.

Table~\ref{table:sba_aht_results} outlines the mean performance for each of the populations and split sizes. In all cases, SBA improves performance over the baseline. For the \textit{medium} SAD splits we also compare to the best result for this population from Generalized Beliefs (see Section~\ref{sec:related_work_aht}) \citep{muglich2022generalized}. Our method improves upon this previous Hanabi AHT state-of-the-art by an average of 2.93 points\footnote{Our baseline (BR) also outperforms Generalised Beliefs because we base our work on the newer and more fine-tuned OBL implementation.}.

We perform two-tailed Monte Carlo permutation tests \citep{dwass1957modified} to estimate whether there is a statistically significant difference between the baseline and SBA. For medium and large split sizes, we find that SBA confers a statistically significant advantage over the baseline to the level $\alpha=0.01$. For small split sizes, we find that the improvement is statistically significant to the level $\alpha=0.05$. This shows that applying SBA while training an AHT agent in this context consistently improves performance, making it essential for AHT when symmetry-based conventions exist within the training population.

We also examine the effect of SBA when applied to OP agents. We train baseline and SBA agents on \textit{medium} OP train/test splits. Intuitively, OP trains agents to be equivariant under symmetries which should render SBA ineffective. Indeed, the baseline agents achieve an average score of 19.27 $\pm$ 0.42 with the held-out partners, while our SBA agents score 19.39 $\pm$ 0.42. We find that this difference is \textit{not statistically significant,} indicating that SBA does not degrade performance even when applied to a population with lower AugImp variance under colour permutation.


\begin{figure}[h]
    \centering
    \includegraphics[width=0.7\linewidth]{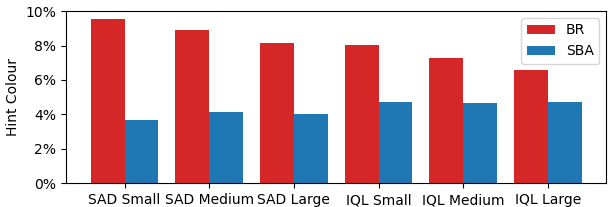}
    \caption{Frequency of the \textit{hint colour} action played. SBA hints colours less often when the training set uses symmetry breaking conventions.}
    \label{fig:sba_actions}
\end{figure}

To gain insight into why SBA improves performance (whenever it does work!), we analyse how often each agent gives \textit{colour hints} (Figure~\ref{fig:sba_actions}). When an SBA agent is trained with SAD and IQL splits, it hints colours significantly less often compared to a baseline agent. Due to random symmetry breaking, hinting colours risks eliciting an unexpected reaction from the teammate, and SBA learns to avoid this. Instead it learns to use other means of hinting, such as using rank. When trained with OP, however, given OP's low AugImp variance (Figure~\ref{fig:augmentation_impact}), it hints colours roughly as often as the baseline.

\subsection{Generalisation to Other Populations}

\begin{wrapfigure}{r}{0.5\textwidth} 
\vspace{-10pt}
\centering
\begin{adjustbox}{width=0.50\textwidth}
\begin{tabular}{llll}
\toprule
Train Set & Agent & OP $\uparrow$ & OBL $\uparrow$ \\
\midrule
\multirow{2}{*}{SAD} & BR  & 15.69 $\pm$ 0.26 & \bfres{4.51}{0.21} \\
                     & SBA (ours) & \bfres{17.72}{0.26} & 3.85 $\pm$ 0.16 \\
\midrule
\multirow{2}{*}{IQL} & BR  & 15.71 $\pm$ 0.24 & \bfres{5.73}{0.24}\\
                     & SBA (ours) & \bfres{16.42}{0.16} & 5.50 $\pm$ 0.20 \\
\bottomrule
\end{tabular}
\end{adjustbox}
\caption{\textit{Symmetry-Breaking Conventions} performance in Hanabi 
for \textit{medium} training set sizes, cooperating with out-of-distribution 
populations. Shown is the standard error of the mean (s.e.m) across 13, 10, 
and 10 training splits respectively.}
\label{table:sba_gen_results}
\vspace{-5pt}
\end{wrapfigure}

The previous experiment examines how SBA affects generalisation to held-out teammates from the same population as those used for training (either SAD or IQL). Here, we investigate whether SBA is effective at creating policies that can transfer to entirely different populations. We take the \textit{medium} split size agents from the previous experiment, and evaluate them when paired with teammates from these different populations.

Table~\ref{table:sba_gen_results} outlines our results. SBA agents trained on SAD exhibit improved transfer performance to the IQL and OP populations. Applying the same significance testing as in the previous experiment, we find that these results are significant to the level $\alpha=0.01$. Similarly for agents trained with IQL partners, we find that SBA confers a statistically significant advantage when playing with OP agents (to the level $\alpha=0.05$). Additional results with all split sizes and OP-trained SBA and baseline agents are available in Appendix~\ref{apx:aht_results}.

We also see that SBA can \textit{harm} performance when transferring to the OBL population. For SBA agents trained with SAD, we find this detrimental effect to be statistically significant to the level $\alpha=0.01$. OBL policies require explicit colour information approximately $65\%$ of the time for cards that they play \citep{hu2021off}, and thus rely on colour hints. SBA agents exhibit much lower frequency of providing these colour hints to the partner (Figure~\ref{fig:sba_actions}). We hypothesize that this is one of the main reasons why SBA harms performance in this case.

\section{Related Work}

\subsection{Ad hoc Teamwork} \label{sec:related_work_aht}

There have been a number of works that address the Ad Hoc Teamwork (AHT) problem. One such technique is Generalised Beliefs that uses belief models to assist with generalisation~\citep{muglich2022generalized}.
Belief models are used to provide latent representations of trajectories to a policy model and assist with search rollouts for action selection.
They show that this leads to improvements in AHT scores in Hanabi.
This technique, however, requires an additional step to train the belief model prior to AHT agent training.

Several approaches achieve generalisation to held-out policies by training a best response to a diverse population~\citep{lowe2017multi, charakorn2020investigating, mckee2022quantifying}, often by training this population using an approach that encourages diversity. Approaches include minimising performance between policies while maximising individual self-play scores \citep{charakorn2022generating, cui2022adversarial, rahman2023generating} and finding minimum coverage sets that span the policy space~\citep{rahman2023minimum, lauffer2023needs}. \citet{canaan2022generating} use hand-crafted rules to find diverse agents that have been trained using genetic algorithms and \citet{yu2023learning} introduce sub-optimal biases into the reward function. While this ensures diversity in the training teammates, training large enough populations can be prohibitively expensive given the range of possible conventions.

Some previous works train an AHT agent using a population that is generated from scratch to be maximally diverse \citep{rahman2023generating, charakorn2022generating} or to approximate the minimum coverage set of possible best-response policies \citep{rahman2023minimum}. Our approach differs from these in that it can be applied to any population, whether pre-existing or generated from scratch. SBA could be used in combination with a population generation-based approach to further increase diversity with a smaller training population size. In the presence of environmental symmetries, this could reduce the time required to train (and load) the AHT training population.

For zero-shot coordination relate work, see Appendix~\ref{apx:zsc}, and for social convention related work, see Appendix~\ref{apx:social_conventions}.

\section{Conclusion}

In this work we have shown that by applying a simple augmentation to the basic AHT learning framework, which we call \textit{symmetry-breaking augmentations}, we can construct agents that are better able to coordinate in the AHT setting with partners they have not seen before. Our method achieves state-of-the-art performance when evaluated with a diverse collection of policies, including SAD policies that were previously unable to collaborate well in cross-play due to their high degree of symmetrical conventional specialisation. We have demonstrated that SBA always improves performance, regardless of training set size, and have defined SBA generally, shown its implementation with deep RL, and provided evidence from experiments in Hanabi that SBA yields robust agents capable of playing well with unfamiliar artificial partners.

One limitation of our approach is that symmetries must exist in both the environment and in teammate strategies, and may require expert knowledge to define. However, we expect that SBA could be applied to partial or imperfect symmetries, or extended to more general augmentations that are not strictly symmetry-based. Methods to automatically detect these symmetries could also be developed, and we leave this for future work.

In our experiments we assumed that a population of training agents is available whose strategies serve as a reasonable prior for the teammates our AHT agent will encounter; nevertheless, SBA could also be applied in settings where this training population is generated from scratch~\citep{lupu2021trajectory, rahman2023generating}.

In future work we will investigate how SBA can be utilised in combination with other AHT improvements, such as search, for instance by shuffling symmetries in search rollouts~\citep{sutton2018reinforcement} to better predict teammate actions. We will also apply SBA to a wider range of Dec-POMDPs, including those with disjoint sets of equivalent states and agents that observe and act in the real world. Given the prevalence of (potentially imperfect) symmetries in the real world, we believe that the key SBA ideas can be used to augment the experiences of real-world agents.

\vspace*{\fill}
\pagebreak

\subsection*{Ethics Statement}

We have carefully considered the ethical implications of our work and have adhered to all relevant institutional, national, and international guidelines. No experiments involving human or animal subjects were conducted, and all data used were either publicly available or obtained with the necessary permissions and anonymised. We welcome further discussion on the ethical aspects of our research.

\subsubsection*{Acknowledgments}
Ravi gratefully acknowledges the funding provided by the Australian Government Research Program (RTP) Scholarship during his time at Adelaide University whilst working on this project. He also acknowledges funding from Autonomous Intelligent Machines and Systems, as well as from Rosebud, during his time at Oxford University.

\bibliography{iclr2025_conference}
\bibliographystyle{iclr2025_conference}

\vspace*{\fill}
\pagebreak

\appendix

\section{Proofs} 

\subsection{Lemma 1} \label{apx:proofs_lemma1}

$J(\pi) = J(\phi(\pi)) \forall \phi \in \Phi, \pi \in \Pi$ 

This Lemma shows that the expected return of a joint-policy $\pi$ is equal to the expected return when $\phi$ is
applied $\pi$.

\begin{proof}
\begin{align}
    J(\pi) &= \mathbb{E}_{\tau_t \sim \pi}V^\pi(\tau_t)\\
    &= \sum_{\tau_t}P(\tau_t|\pi)\sum_{a_t}\pi(a_t|\tau_t)\sum_{r_t}\mathcal{R}(r_t|s_t, a_t) \nonumber \\
    &~~~~~\Bigg(\Bigg. r_t + \gamma\sum_{s_{t+1}}\mathcal{T}(s_{t+1}|s_t, a_t)\sum_{o_{t+1}}\mathcal{U}(o_{t+1}|s_{t+1}, a_t)V^{\pi}(\tau_t \oplus (a_t, o_{t+1})) \Bigg)\Bigg. \\
    &= \sum_{\tau_t}P(\tau_t|\phi(\phi^{-1}(\pi)))\sum_{a_t}\pi(\phi(\phi^{-1}(a_t))|\phi(\phi^{-1}(\tau_t)))\sum_{r_t}\mathcal{R}(r_t|s_t, a_t) \nonumber \\
    &~~~~~\Bigg(\Bigg. r_t + \gamma\sum_{s_{t+1}}\mathcal{T}(s_{t+1}|s_t, a_t)\sum_{o_{t+1}}\mathcal{U}(o_{t+1}|s_{t+1}, a_t)V^{\phi(\phi^{-1}(\pi))}(\tau_t \oplus (a_t, o_{t+1})) \Bigg)\Bigg. \\
    &~\text{Since }\phi \text{ is an automorphism}. \nonumber \\
    &= \sum_{\phi(\tau_t)}P(\phi(\tau_t)|\phi(\phi^{-1}(\phi(\pi)))\sum_{\phi(a_t)}\pi(\phi(\phi^{-1}(\phi(a_t)))|\phi(\phi^{-1}(\phi(\tau_t)))\sum_{\phi(r_t)}\mathcal{R}(\phi(r_t)|\phi(s_t), \phi(a_t)) \nonumber \\
    &~~~~~\Bigg(\Bigg. \phi(r_t) + \gamma\sum_{\phi(s_{t+1})}\mathcal{T}(\phi(s_{t+1})|\phi(s_t), \phi(a_t))\sum_{\phi(o_{t+1})}\mathcal{U}(\phi(o_{t+1})|\phi(s_{t+1}), \phi(a_t)) \nonumber \\
    &~~~~~V^{\phi(\phi^{-1}(\phi(\pi)))}(\phi(\tau_t) \oplus (\phi(a_t), \phi(o_{t+1}))) \Bigg)\Bigg. \\
    &= \sum_{\tau_t}P(\tau_t|\phi(\pi))\sum_{a_t}\pi(\phi(a_t)|\phi(\tau_t))\sum_{r_t}\mathcal{R}(r_t|s_t, a_t) \nonumber \\
    &~~~~~\Bigg(\Bigg. r_t + \gamma\sum_{s_{t+1}}\mathcal{T}(s_{t+1}|s_t, a_t)\sum_{o_{t+1}}\mathcal{U}(o_{t+1}|s_{t+1}, a_t)V^{\phi(\pi)}(\tau_t \oplus (a_t, o_{t+1})) \Bigg)\Bigg. \\
    &= \mathbb{E}_{\tau_t \sim \phi(\pi)}V^{\phi(\pi)}(\tau_t)\\
    &= J(\phi(\pi))
\end{align}
\end{proof}

\vspace*{\fill}
\pagebreak

\subsection{Proposition 1} 

The expected AHT return when $\phi$ is applied to $\pi$ is equal to the expected AHT return when the inverse $\phi^{-1}$ is applied to each of the $\pi^i_j$ teammate policies. \label{apx:proofs_prop1}

\begin{proof}
\begin{align} \label{eqn:symmetry_inverse_equality}
    J&_{AHT}(\pi_A, \phi^{-1}(\pi_j)) \nonumber \\
        &= \frac{1}{n} \big( J(\pi^1_A,\phi^{-1}(\pi^2_j), \dots, \phi^{-1}(\pi^n_j)) + \nonumber \\
        &~~~~~~~~\dots + J(\phi^{-1}(\pi^1_j), \dots, \phi^{-1}(\pi^{n-1}_j), \pi^n_A) \big) \nonumber \\
        &= \frac{1}{n} \big( J(\phi(\pi^1_A),\phi(\phi^{-1}(\pi^2_j)), \dots, \phi(\phi^{-1}(\pi^n_j))) + \nonumber \\
        &~~~~~~~~\dots + J(\phi(\phi^{-1}(\pi^1_j)), \dots, \phi(\phi^{-1}(\pi^{n-1}_j)), \phi(\pi^n_A)) \big)  \nonumber \\
        &= \frac{1}{n} \big( J(\phi(\pi^1_A),\pi^2_j, \dots, \pi^n_j) + \nonumber \\
        &~~~~~~~~\dots + J(\pi^1_j, \dots, \pi^{n-1}_j, \phi(\pi^n_A)) \big) \nonumber \\
        &= J_{AHT}(\phi(\pi_A), \pi_j) 
\end{align}
\end{proof}

\section{Iterated Lever Coordination Game Details} \label{apx:lever_experiment_setup}

To emphasise the necessity of augmenting a training policy population with symmetry-breaking augmentation, we have created the iterated lever-coordination game. This game underscores the importance of exposing AHT agents to conventions not initially present in the training population to facilitate generalisation to a broader range of conventions at test-time. The neural network employed in this experiment is a basic 2-layer fully connected network with one hidden layer, utilising the sigmoid function as the activation function. The training process takes place on a CPU with a single thread. We present the crucial hyper-parameters in Table~\ref{table:lever_game_details}.

\begin{table}[h]
\centering
\begin{tabular}{ll}
\toprule
Hyper-parameters & Value \\
\midrule
\texttt{\# Network} & \\
\texttt{hidden size} & \texttt{20} \\
\texttt{activation} & \texttt{sigmoid} \\ 
\texttt{layers} & \texttt{2} \\       
\midrule
\texttt{\# Optimisation} & \\
\texttt{optimiser} & \texttt{Adam} \\
\texttt{lr} & \texttt{0.05} \\
\texttt{eps} & \texttt{0.9} \\
\texttt{batchsize} & \texttt{10} \\
\midrule
\texttt{\# Training} & \\
\texttt{epochs} & \texttt{1000} \\
\texttt{num runs} & \texttt{30} \\
\bottomrule
\end{tabular}
\caption{Hyper-Paramaters for \textit{iterated lever coordination game} Reinforcment Learning.}
\label{table:lever_game_details}
\end{table}

\section{Hanabi Rules} \label{apx:hanabi_rules}

Hanabi is a co-operative card game where players work together to create five colour-coded stacks of cards, each stack arranged in ascending rank from one to five. The deck contains exactly fifty cards: five colours (red, yellow, green, blue, and white), each composed of three copies of rank one, two copies of ranks two, three, and four, and a single copy of rank five. The game is played with two to five players. If there are two or three players, each starts with five cards; if there are four or five players, each starts with four. Players hold their cards facing away from themselves so they can see everyone else’s hand, but not their own. The group shares eight information tokens and three fuse tokens. If the group ever loses all three fuse tokens, the game ends immediately and the final score is zero.

Each turn, a player must choose one of three actions. The first action is to \textbf{give information} to a teammate by spending one information token. This clue must focus on a single rank or a single colour, and the clue-giver must indicate every card in the teammate’s hand that matches that choice. The second action is to \textbf{discard} a card from hand, which returns one information token to the pool (unless the group already has the maximum of eight). A new card is drawn from the deck to replace any discarded card if the deck has not yet been exhausted. The third action is to \textbf{play} a card from hand, attempting to place it on the appropriate stack. Each colour stack must begin with a rank one, followed by two, three, four, and five in ascending order. A card that is played correctly is added to its colour stack, and if that card is a rank five, the team gains one information token (up to a maximum of eight). If a played card cannot legally be placed (for example, it is the wrong rank for its colour stack), a bomb is triggered and one fuse token is removed. 

Once the deck is empty, each player takes one final turn. The score is the total number of successfully placed cards across all colours, with a maximum of twenty-five if every card is played in perfect sequence. Communication is strictly limited to the “give information” action, so effective co-operation relies on careful deduction and subtle signalling to avoid bombs and achieve the highest possible score.

\section{Experiment Details for Hanabi} \label{apx:hanabi_experiment_setup}

\subsection{Reinforcement Learning} 

\begin{figure}[h]
    \centering      
    \includegraphics[width=0.6\linewidth]{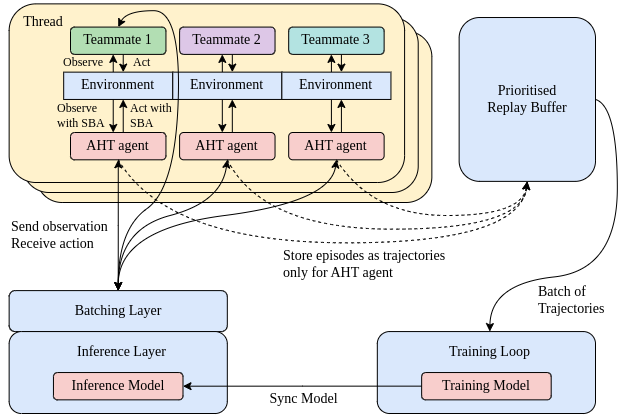}
    \caption{Illustration of RL training setup for AHT using SBA. Some arrows linking \textit{player 1} and \textit{batching layer} are omitted for legibility. Note that the experience of the teammates are not stored in the replay buffer, or used to update the model in the \textit{training loop}.}
    \label{fig:training_architecture}
\end{figure}

We employ a highly scalable training architecture illustrated in Figure~\ref{fig:training_architecture}, built upon the framework implemented in the Off-Belief Learning Github Repository~\citep{hu2021off}. This architecture features multiple parallel thread workers responsible for managing interactions across various Hanabi environments and the agents operating within each environment. Each actor initiates multiple inference calls to neural networks at every time step, with each inference call executed on GPUs. When a player initiates an inference call, the worker thread promptly proceeds to the next agent in the thread, facilitating the simultaneous execution of multiple games and agents on a single worker thread. Inferences invoked by different players are batched together and processed on the GPU in parallel. This approach enables the concurrent execution of a substantial number of games and environments, generating a significant volume of data for training purposes.

At each time step, players collect observations, actions, and rewards, and aggregate them into episodes. These trajectories are padded to 80 time steps and stored in a priority replay buffer. The training loop, operating independently of the aforementioned worker threads, continually samples transitions from the buffer, using them to update the model. After every 10 gradient update steps, the new model synchronizes with all the models conducting inference on GPUs.

We extended the training architecture to accommodate Ad Hoc Teamwork (AHT) agents. In each game within every worker thread, one agent acts as our AHT agent, learning from the experience, while the other agent sends inference requests to a frozen set of pre-trained neural network weights. Each AHT agent interacts with a distinct pre-trained model. When agents collect observations, actions, and rewards for storage in the replay buffer, only the AHT agent's experience is used for optimization; teammates' experiences are discarded. At the episode's start, a different policy augmentation permutation is chosen, maintaining an unchanged teammate in the environment. To ensure an even distribution of teammate policies, the number of games is selected to perfectly divide the number of training agents, ensuring experiences are evenly distributed across teammate policies.

The architecture adheres to what was, at one point in time, a state-of-the-art model—Recurrent Replay Distributed Deep Q-Networks (R2D2)~\citep{kapturowski2018recurrent}—which incorporates best practices such as double-DQN~\citep{van2016deep}, dueling network architecture~\citep{wang2016dueling}, prioritized experience replay~\citep{schaul2015prioritized}, distributed training with parallel environments~\citep{horgan2018distributed}, and a recurrent network to handle partial observability~\citep{graves2012long}. In all our experiments, we execute games and players on 23 worker threads, with 80 games per thread, and allocate 1 thread for training. For inferences, we employ three Nvidia GeForce RTX 3090 GPUs and one for training, and the worker threads are executed on an AMD Ryzen Threadripper 1920X 12-Core CPU. Our essential hyper-parameters are detailed in Table~\ref{table:hanabi_hyperparameters}.

\begin{table}[h]
\centering
\begin{tabular}{ll}
\toprule
Hyper-parameters & Value \\
\midrule
\texttt{\# replay buffer related} & \\
\texttt{burn-in frames} & \texttt{10,000} \\
\texttt{replay buffer size} & \texttt{100,000} \\ 
\texttt{priority exponent} & \texttt{0.9} \\       
\texttt{priority weight} & \texttt{0.6} \\       
\texttt{max trajectory length} & \texttt{80} \\       
\midrule
\texttt{\# Optimisation} & \\
\texttt{optimiser} & \texttt{Adam} \\
\texttt{lr} & \texttt{6.25e-05} \\
\texttt{eps} & \texttt{1.5e-05} \\
\texttt{grad clip} & \texttt{5} \\
\texttt{batchsize} & \texttt{128} \\
\midrule
\texttt{\# Q learning} & \\
\texttt{n step} & \texttt{3} \\
\texttt{discount factor} & \texttt{0.999} \\
\texttt{target network sync interval} & \texttt{2500} \\
\texttt{exploration $\epsilon$} & \texttt{$\epsilon_0 \dots \epsilon_n, \text{where } \epsilon_i = 0.1^{1+7_i/(n-1)}, n=80$} \\
\bottomrule
\end{tabular}
\caption{Hyper-Paramaters for \textit{Hanabi} Reinforcment Learning.}
\label{table:hanabi_hyperparameters}
\end{table}

\subsection{Hanabi Training Population Splits} \label{apx:population_splits}

To gain a more accurate understanding of SBA's performance when learning how to generalize a held-out sets of evaluation policies, it is useful to observe how much generalisation is affected by the size of the training set. Since SBA combinatorially increases the number of partners encountered during training, a performance improvement should still be observed when there is only one policy in the training set. As there is a varying number of pre-trained policies in each population, specifically 13 simplified action decoder (SAD), 12 independent Q-learning (IQL), and 12 other-play (OP), we have different split sizes for \textit{small}, \textit{medium}, and \textit{large} splits. Table~\ref{table:train_test_splits} provides the exact breakdown of these splits.

\begin{table}[h]
\centering
\begin{tabular}{llll}
\toprule
Population & \textit{small} splits & \textit{medium} splits & \textit{large} splits \\
\midrule
\texttt{SAD} & \texttt{1 train/12 test} & \texttt{6 train/7 test} & \texttt{11 train/2 test} \\
\texttt{IQL} & \texttt{1 train/11 test} & \texttt{6 train/6 test} & \texttt{10 train/2 test} \\
\texttt{OP} & \texttt{1 train/11 test} & \texttt{6 train/6 test} & \texttt{10 train/2 test} \\
\bottomrule
\end{tabular}
\caption{Breakdown of the train and test splits we use for Hanabi policy populations.}
\label{table:train_test_splits}
\end{table}

\section{Augmentation Impact Breakdown} \label{apx:augmentation_impact}

Training an AHT agent in large Dec-POMDPs can be expensive, so it's important to determine whether an augmentation technique will meaningfully diversify a population before training commences. SBA is a technique that will only change the policies in the training population if they rely on symmetry-breaking conventions. Therefore, before training, we can use Augmentation Impact (AugImp) (Equation~\ref{eqn:augmentation_difference}) to assess how much a given augmentation (in our case, SBA) will diversify the policies in a population.

\begin{figure}[H]
    \centering    
    \vspace{-10pt}
    \includegraphics[width=0.9\linewidth]{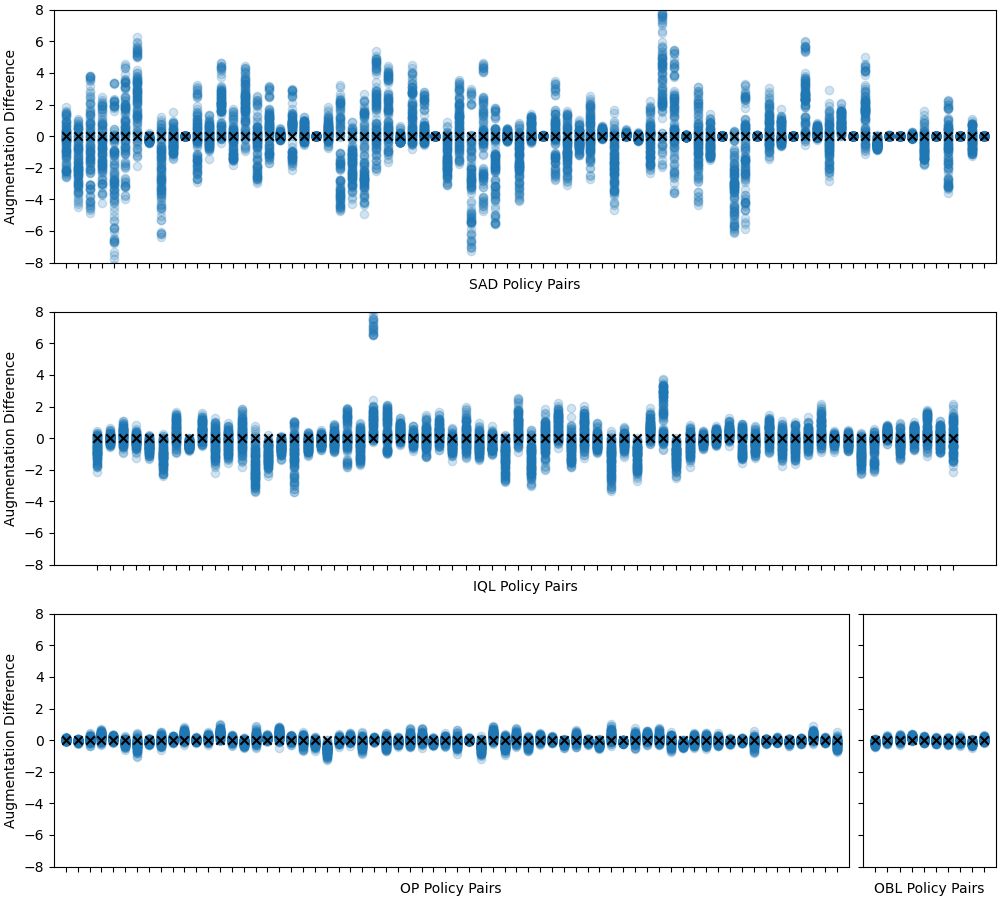}
    \caption{\textit{Augmentation Difference} (Equation~\ref{eqn:augmentation_difference}) scores for all pairs of SAD, IQL, OP, and OBL. Each column of blue contains the score differences between two policy pairs before and after applying an augmentation. The black x's represent the original non-augmented policies. It's clear that the SBA diversifies SAD and IQL populations much more than OP and OBL.}
    \label{fig:sba_all_score_differences}
\end{figure}

The AugImp calculates the absolute difference across all pairs of agents and across all permutations. This metric combines all the information for a population into a single value, but the information about the max and min values is lost. To obtain a more thorough overview of the augmentations, we examine all Augmentation Differences (AD) individually for each policy pair. AD represents the Hanabi score difference between two agents from a population before and after the permutation has been applied to one agent, and it is defined as:
\begin{equation} \label{eqn:augmentation_difference}
    AD(\pi_1, \pi_2, \phi) = J_{\text{XP}}(\pi_1, \pi_2) - J_{\text{XP}}(\phi(\pi_1), \pi_2).
\end{equation}
In Figure~\ref{fig:sba_all_score_differences}, we can observe the complete breakdown of all \textit{Augmentation Differences} (AD) for all four Hanabi populations: SAD, IQL, OP, and OBL. The plot illustrates how different the augmentation scores are compared to no augmentation (depicted as a black 'x'). As expected, the plot demonstrates that SAD and IQL policy populations have a much larger spread than OP and OBL, with SAD frequently reaching an Augmentation Difference of $\pm$ 8 points. Interestingly, there exists one IQL pair that works together particularly well when a specific color permutation is applied.

\section{AHT Results} \label{apx:aht_results}

\subsection{SAD Ad Hoc Teamwork} \label{apx:sad_aht_results}

\begin{table}[H]
\centering
\begin{adjustbox}{width=1\textwidth}
\begin{tabular}{llllllllll}
\toprule
&& \multicolumn{2}{c}{SAD (eval)} & \multicolumn{2}{c}{w/ IQL} & \multicolumn{2}{c}{w/ OP} & \multicolumn{2}{c}{w/ OBL}\\
Split & Agent & Score $\uparrow$ & Bombout $\downarrow$ & Score $\uparrow$ & Bombout $\downarrow$ & Score $\uparrow$ & Bombout $\downarrow$  & Score $\uparrow$ & Bombout $\downarrow$ \\
\midrule
\multirow{2}{*}{small} & BR         & 8.15 $\pm$ 1.28     & 0.60 $\pm$ 0.03    & 10.78 $\pm$ 1.34 & 0.46 $\pm$ 0.06 & 10.85 $\pm$ 1.50    & 0.47 $\pm$ 0.07    & \bfres{3.77}{0.39} & 0.70 $\pm$ 0.01 \\
                      & SBA (ours) & \bfres{9.12}{1.42}  & \bfres{0.57}{0.03} & \bfres{11.53}{1.39} & \bfres{0.44}{0.06} & \bfres{12.19}{1.59} & \bfres{0.41}{0.07} & 3.65 $\pm$ 0.34   & 0.71 $\pm$ 0.01 \\
\midrule
\multirow{3}{*}{medium}  & Gen. Belief& 12.36 $\pm$ 0.96    & -                  & - & - & -                   & -                  & -                 & - \\
                      & BR         & 13.09 $\pm$ 0.49    & 0.39 $\pm$ 0.04    & 15.22 $\pm$ 0.25 & 0.27 $\pm$ 0.01 & 15.69 $\pm$ 0.26    & 0.26 $\pm$ 0.01    & \bfres{4.51}{0.21} & \bfres{0.65}{0.01} \\
                      & SBA (ours) & \bfres{15.40}{0.49} & \bfres{0.28}{0.04} & \bfres{16.71}{0.22} & \bfres{0.20}{0.01} & \bfres{17.72}{0.26} & \bfres{0.17}{0.01} & 3.85 $\pm$ 0.16   & 0.70 $\pm$ 0.01 \\
\midrule
\multirow{2}{*}{large} & BR         & 14.69 $\pm$ 1.05    & 0.31 $\pm$ 0.05    & 16.61 $\pm$ 0.12 & 0.21 $\pm$ 0.01 & 16.78 $\pm$ 0.22    & 0.21 $\pm$ 0.01    & \bfres{4.56}{0.22} & 0.64 $\pm$ 0.02 \\
                      & SBA (ours) & \bfres{16.34}{1.29} & \bfres{0.24}{0.06} & \bfres{17.40}{0.07} & \bfres{0.17}{0.00} & \bfres{17.81}{0.21} & \bfres{0.16}{0.01} & 4.36 $\pm$ 0.20   & 0.65 $\pm$ 0.02 \\
\bottomrule
\end{tabular}
\end{adjustbox}
\caption{Mean scores and bombout rates for BR and SBA in Hanabi with a SAD population. Models are trained with 1 train and 12 test policies, 6-7, and 11-2. Compared are the reported 6-7 split scores for Generalized Beliefs (gen. belief)~\citep{muglich2022generalized} (gen. belief). Shown is the standard error of the mean (s.e.m) across 13, 10, and 10 training splits respectively. An AHT agent trained on a SAD policies significantly improves performance over the baseline.}
\label{table:sad_sba_results}
\end{table}

\begin{table}[h]
\centering
\begin{tabular}{lllll}
\toprule
Splits   &SAD (eval)  &IQL   &OP    &OBL \\
\midrule
1-12      &\textbf{0.021} &\textbf{0.019} &\textbf{0.002} &0.648 \\
6-7       &\textbf{0.002} &\textbf{0.002} &\textbf{0.002} &\textbf{0.006}\\
11-2      &\textbf{0.002} &\textbf{0.002} &\textbf{0.002} &0.438\\
\bottomrule
\end{tabular}
\caption{Monte carlo paired permutation test comparing SBA to BR trained on SAD across different splits, and evaluating with different teammates. 100k samples are taken. Calculated is the two-sided p-value.}
\label{table:sad_sba_significance}
\end{table}

Table~\ref{table:sad_sba_results} showcases the performance results of SBA and baseline (BR) agents, both trained with SAD training populations of varying sizes: small, medium, and large. The robustness of each AHT agent is assessed by evaluating them with a held-out set of SAD agents, while their generalization is tested through cooperation with three distinct algorithms—namely, IQL, OP, and OBL—that were absent during training. In addition to the Hanabi scores, the bombout rate is presented, where lower rates indicate superior performance. The displayed standard error of the mean reflects the error across the training splits.

Notably, when the SBA agent is evaluated in coordination with other held-out SAD agents, as well as agents from IQL and OP populations, there is a significant improvement in scores across all scenarios. The most substantial improvement percentage is observed when SBA is trained with a medium-sized training set, resulting in up to a 17\% score improvement. Remarkably, even when trained with just one agent in the training set, SBA still demonstrates performance enhancement, highlighting the efficacy of symmetry-breaking augmentations.

Interestingly, when SAD agents coordinate with OBL policies, the performance declines. This can be attributed to the fact that, in approximately 60\% of instances, OBL plays a card only when it knows both the color and rank information. Since SBA tends to provide color hints less frequently, it fails to furnish OBL with sufficient information, resulting in a reduced frequency of card plays and, consequently, diminished performance.

In determining statistical significance, we employ a Paired Monte-Carlo Permutation test, and the corresponding results are presented in Table~\ref{table:sad_sba_significance}. Notably, SBA demonstrates a significant performance enhancement when assessed under SAD, IQL, and OP training policies. Conversely, its performance takes a noticeable dip when the AHT agent engages with OBL, particularly for medium-sized training sets. However, for both small and large training sets, the outcomes remain inconclusive.

\subsection{IQL Ad Hoc Teamwork} \label{apx:iql_aht_results}

\begin{table}[H]
\centering
\begin{adjustbox}{width=1\textwidth}
\begin{tabular}{llllllllll}
\toprule
&& \multicolumn{2}{c}{IQL (eval)} & \multicolumn{2}{c}{w/ SAD} & \multicolumn{2}{c}{w/ OP} & \multicolumn{2}{c}{w/ OBL}\\
Split & Agent & Score $\uparrow$ & Bombout $\downarrow$ & Score $\uparrow$ & Bombout $\downarrow$ & Score $\uparrow$ & Bombout $\downarrow$  & Score $\uparrow$ & Bombout $\downarrow$\\
\midrule
\multirow{2}{*}{1-11} & BR &  11.52 ± 1.08 &  0.41 ± 0.05 &  8.49 ± 1.19 &  0.59 ± 0.05 &  11.52 ± 1.34 &  0.43 ± 0.06 &  5.19 ± 0.26 &  0.58 ± 0.02 \\
                      & SBA (ours) &  11.84 ± 0.99 &  0.40 ± 0.04 &  8.32 ± 0.96 &  0.59 ± 0.04 &  11.91 ± 1.08 &  0.40 ± 0.05 &  4.13 ± 0.33 &  0.66 ± 0.03 \\
\midrule
\multirow{2}{*}{6-6}  & BR         & 15.04 $\pm$ 0.37 & 0.27 $\pm$ 0.02 & 13.23 $\pm$ 0.15 & 0.38 $\pm$ 0.01 & 15.71 $\pm$ 0.24 & 0.25 $\pm$ 0.01 & 5.73 $\pm$ 0.24 & 0.57 $\pm$ 0.02\\
                      & SBA (ours) & \bfres{16.08}{0.42} & \bfres{0.21}{0.02} & 13.70 $\pm$ 0.19 & 0.34 $\pm$ 0.01 & \bfres{16.42}{0.16} & \bfres{0.20}{0.01} & 5.50 $\pm$ 0.20 & 0.56 $\pm$ 0.01 \\
\midrule
\multirow{2}{*}{10-2}  & BR &  15.34 ± 0.80 &  0.25 ± 0.03 &  13.81 ± 0.11 &  0.34 ± 0.00 &  16.06 ± 0.17 &  0.22 ± 0.01 &  5.97 ± 0.16 &  0.54 ± 0.01 \\
                       & SBA (ours)&  \bfres{15.95}{0.71}&  \bfres{0.21}{0.03} &  14.08 ± 0.16 &  0.33 ± 0.01 &  \bfres{16.72}{0.11} &  \bfres{0.20}{0.01} &  5.93 ± 0.16 &  0.55 ± 0.01 \\

\bottomrule
\end{tabular}
\end{adjustbox}
\caption{Mean scores and bombout rates for BR and SBA in Hanabi with a IQL population. Models are trained with 1 train and 12 test policies, 6-7, and 11-2. Shown is the standard error of the mean across 13, 10, and 10 training splits respectively.}
\label{table:iql_sba_results}
\end{table}

\begin{table}[h]
\centering
\begin{tabular}{lllll}
\toprule
Splits   &IQL (eval)  &SAD   &OP    &OBL \\
\midrule
1-11      &0.187 &0.713 &0.156 &0.117 \\
6-6       &\textbf{0.004} &0.156 &\textbf{0.021} &0.330 \\
10-2      &\textbf{0.004} &0.235 &\textbf{0.031} &0.284 \\
\bottomrule
\end{tabular}
\caption{Monte carlo paired permutation test comparing SBA to BR trained on IQL across different splits, and evaluating with different teammates. 100k samples are taken. Calculated is the two-sided p-value.}
\label{table:iql_sba_significance}
\end{table}

Table~\ref{table:iql_sba_results} presents the performance results of SBA and baseline (BR) agents, both trained with IQL training populations of varying sizes: small, medium, and large. The robustness of each AHT agent is assessed by evaluating them with a held-out set of IQL agents, while their generalization is tested through cooperation with three distinct algorithms—specifically, SAD, OP, and OBL—that were absent during training. In addition to the Hanabi scores, the bombout rate is provided, with lower rates indicating superior performance. The displayed standard error of the mean reflects the error across the training splits.

In contrast to when SBA is trained with a SAD population, its performance does not exhibit the same level of strong improvement when trained with IQL. This is expected due to the lower Augmentation Impact score it receives (Section~\ref{sec:teammate_selection}). Nevertheless, this AHT agent still demonstrates statistically significant performance on an IQL evaluation set for medium and large-sized training sets, as well as when evaluated with OP policies for medium and large training sets. However, when SBA is trained with a single IQL training partner, it never shows statistically significant performance over the baseline (BR). This is likely because the conventions present in a single IQL agent are not strong enough for SBA to meaningfully create a diverse training population.

Furthermore, as shown in Table~\ref{table:iql_sba_significance}, none of these SBA agents achieve statistically significant improvements over the baseline when playing with SAD agents. This is likely attributed to the fact that SAD conventions are much stronger than IQL, and learning to adapt to IQL conventions alone is not sufficient to adequately adjust to SAD.

\subsection{OP Ad Hoc Teamwork} \label{apx:op_aht_results}

\begin{table}[H]
\centering
\begin{adjustbox}{width=1\textwidth}
\begin{tabular}{llllllllll}
\toprule
&& \multicolumn{2}{c}{w/ OP (eval)} & \multicolumn{2}{c}{w/ SAD} & \multicolumn{2}{c}{w/ IQL} & \multicolumn{2}{c}{w/ OBL}\\
Split & Agent & Score $\uparrow$ & Bombout $\downarrow$ & Score $\uparrow$ & Bombout $\downarrow$ & Score $\uparrow$ & Bombout $\downarrow$  & Score $\uparrow$ & Bombout $\downarrow$\\
\midrule
\multirow{2}{*}{6-6}  & BR         & 19.27 $\pm$ 0.42 & 0.14 $\pm$ 0.02 & 12.58 $\pm$ 0.18 & 0.40 $\pm$ 0.01 & 15.44 $\pm$ 0.07 & 0.25 $\pm$ 0.00 & 7.27 $\pm$ 0.26 & 0.54 $\pm$ 0.01 \\
                      & SBA (ours)  & 19.39 $\pm$ 0.42 & 0.14 $\pm$ 0.02 & 12.51 $\pm$ 0.24 & 0.40 $\pm$ 0.01 & 15.52 $\pm$ 0.13 & 0.24 $\pm$ 0.01 & 7.44 $\pm$ 0.39 & 0.53 $\pm$ 0.02 \\
\bottomrule
\end{tabular}
\end{adjustbox}
\caption{Mean scores and bombout rates for BR and SBA in Hanabi with a OP population. Models are trained with 1 train and 12 test policies, 6-7, and 11-2. Shown is the standard error of the mean (s.e.m) across 13, 10, and 10 training splits respectively.}
\label{table:op_sba_results}
\end{table}

\begin{table}[h]
\centering
\begin{tabular}{lllll}
\toprule
Splits   &OP (eval)  &SAD   &IQL    &OBL \\
\midrule
6-6       &0.164 &0.641 &0.408 &0.629 \\
\bottomrule
\end{tabular}
\caption{Monte carlo paired permutation test comparing SBA to BR trained on OP across different splits, and evaluating with different teammates. 100k samples are taken. Calculated is the two-sided p-value.}
\label{table:op_sba_signficance}
\end{table}

Table~\ref{table:op_sba_results} presents the performance results of both SBA and baseline (BR) agents, trained with a medium-sized OP training population. The robustness of each AHT agent is evaluated by assessing them with a held-out set of OP agents. Generalization is tested by cooperation with three distinct algorithms—specifically, SAD, IQL, and OBL—that were absent during training. In addition to the Hanabi scores, the bombout rate is provided, where lower rates indicate superior performance. The displayed standard error of the mean (s.e.m) represents the error across the training splits.

Given that an agent trained with Other-Play aims to avoid symmetry-breaking conventions, as supported by its low Augmentation Impact score in Section~\ref{sec:teammate_selection}, augmenting an OP policy with SBA is anticipated to have minimal impact on AHT performance. The results presented in this section substantiate this assertion, demonstrating that SBA does not exhibit any performance improvement over the baseline. In certain cases, due to variance, it even performs slightly worse than the baseline. This observation is further corroborated by the Monte-Carlo Permutation Tests in Table~\ref{table:op_sba_signficance}, where SBA does not show any significantly improved performance, with $p$-values ranging from 0.16 to 0.6.

\section{Related Work Extended}

\subsection{Zero-shot Coordination} \label{apx:zsc}

In zero-shot coordination (ZSC), agents must coordinate with new teammates that are also optimised for ZSC. Example solution approaches to this problem address symmetries in conventions by training agents in self-play with symmetry-equivalent versions of themselves~\citep{hu2020other, treutlein2021new}, incorporating symmetrization into the network architecture ~\citep{muglich2022equivariant}, or using belief models to find optimal grounded policies that assume all previous actions were taken by the uniform random policy~\citep{hu2021off}. While these techniques achieve high scores in ZSC, they are designed to avoid use of specialised conventions, and fail to coordinate with policies that do use these conventions.

\subsection{Social Conventions} \label{apx:social_conventions}

Like humans who use social conventions to facilitate coordination~\citep{hechter2001social, lewis2008convention}, artificial learning agents are also known to exploit conventions to cooperate~\citep{airiau2014emergence}.
The issue, however, is that in many collaborative settings there are multiple optimal strategies under self-play~\citep{tesauro1994td}, but no guarantee that two independently trained agents will converge to policies with compatible conventions ~\citep{foerster2019bayesian, hu2019simplified, hu2020other}.

Solutions have been proposed to encourage learning agents to better converge to test-time conventions. 
Such as revealing test-time observations during training~\citep{lerer2019learning}, learning with human behavioral-cloned models to better coordinate with real humans~\citep{carroll2019utility}, exploiting the similarity between action and observation features to take human-like actions~\citep{ma2023learning}, and by training with teammates that have hidden biases to better coordinate with sub-optimal humans~\citep{yu2023learning}.
While interesting directions, these methods make assumptions about what conventions the test-time policies will use, whereas our approach exploits and diversifies the conventions that already exist within a training population.

\end{document}